\documentclass[11pt,a4paper]{article}
\usepackage[hyperref]{acl2021}
\usepackage{times}
\usepackage{latexsym}
\usepackage{subcaption,booktabs}
\usepackage{mathtools, nccmath, multicol}
\usepackage{comment}

\usepackage{adjustbox}
\usepackage{todonotes}
\usepackage{multirow}
\usepackage{graphicx}
\usepackage{subcaption}
\usepackage{makecell}
\usepackage{semicruch}
\usepackage{float}
\usepackage{hyperref}

\usepackage{microtype}
\usepackage{amsmath}
\usepackage{color, colortbl}
\usepackage{xcolor}
\newcommand{\gray}[1]{\textcolor{gray}{#1}}
\newcommand{\cM}{\mathcal{M}}
\newcommand{\cL}{L}
\newcommand{\cH}{R}
\newcommand{\dL}{\mathcal{D}_{L}}
\newcommand{\cP}{\mathcal{P}}
\newcommand{\tdL}{\mathcal{D}_{L_R}}
\newcommand{\dH}{\mathcal{D}_{R}}
\newcommand{\dHL}{B_{R\rightarrow L}}
\newcommand{\dLH}{B_{L\rightarrow R}}
\newcommand{\tdHL}{B_{R\rightarrow L_R}}
\newcommand{\tdLH}{B_{L_R\rightarrow R}}
\newcommand{\our}{RelateLM}
\newcommand{\rdlfull}{Related Prominent Language}
\newcommand{\rdl}{RPL}

\newcommand{\hbert}{Hi-BERT}
\newcommand{\ebert}{BERT}
\newcommand{\mbert}{mBERT}
\newcommand{\Translit}{Translit}

\definecolor{Gray}{gray}{0.9}

\aclfinalcopy 

\newcommand{\reffig}[1]{Figure \ref{#1}}

\newcommand{\refsec}[1]{Section \ref{#1}}

\title{Exploiting Language Relatedness for Low Web-Resource Language Model Adaptation: An Indic Languages Study 
}

\author{Yash Khemchandani \And 
        Sarvesh Mehtani \And
        Vaidehi Patil \And 
        Abhijeet Awasthi \And
        Partha Talukdar \And 
        Sunita Sarawagi}
        
\author{Yash Khemchandani\textsuperscript{$1$}\thanks{\hspace{3pt}Authors contributed equally} \hspace{.3cm}   Sarvesh Mehtani\textsuperscript{$1$}\footnotemark[1] \hspace{.3cm}
Vaidehi Patil\textsuperscript{$1$}
\\ 
\textbf{ 
 Abhijeet Awasthi\textsuperscript{$1$}
 \hspace{.3cm}
Partha Talukdar\textsuperscript{$2$}
\hspace{.3cm} Sunita Sarawagi\textsuperscript{$1$} }\\
  \textsuperscript{$1$}Indian Institute of Technology Bombay, India\\
  \textsuperscript{$2$}Google Research, India\\
  \texttt{\{yashkhem,smehtani,awasthi,sunita\}@cse.iitb.ac.in}\\
  \texttt{vaidehipatil@ee.iitb.ac.in}, \hspace{.3cm} \texttt{partha@google.com} 
  }

\date{}

\begin{document}
\maketitle
\begin{abstract}

Recent research in multilingual language models (LM) has demonstrated their ability to effectively handle multiple languages in a single model. This holds promise for low web-resource languages (LRL) as multilingual models can enable transfer of supervision from high resource languages to LRLs. However, incorporating a new language in an LM still remains a challenge, particularly for languages with limited corpora and in unseen scripts. In this paper we argue that \emph{relatedness} among 
languages in a language family may be exploited to overcome some of the corpora limitations of LRLs, and propose \our{}. We focus on Indian languages, and exploit relatedness along two dimensions: (1) \emph{script} (since 
many Indic scripts originated from the Brahmic script), and (2) \emph{sentence structure}. \our{} uses transliteration to convert the unseen script of limited LRL text into the script of a  \rdlfull{} (\rdl{}) (Hindi in our case). While exploiting similar sentence structures, \our{} utilizes readily available bilingual dictionaries to pseudo translate \rdl{} text into LRL corpora. 
Experiments on  multiple real-world benchmark datasets provide validation to our hypothesis that using a related language as pivot, along with transliteration and pseudo translation based data augmentation, can be an effective way to adapt LMs for LRLs, rather than direct training or pivoting through English. 


\end{abstract}

\section{Introduction}
\label{sec:intro}

\begin{figure}[t!]
\includegraphics[width=7.5cm]{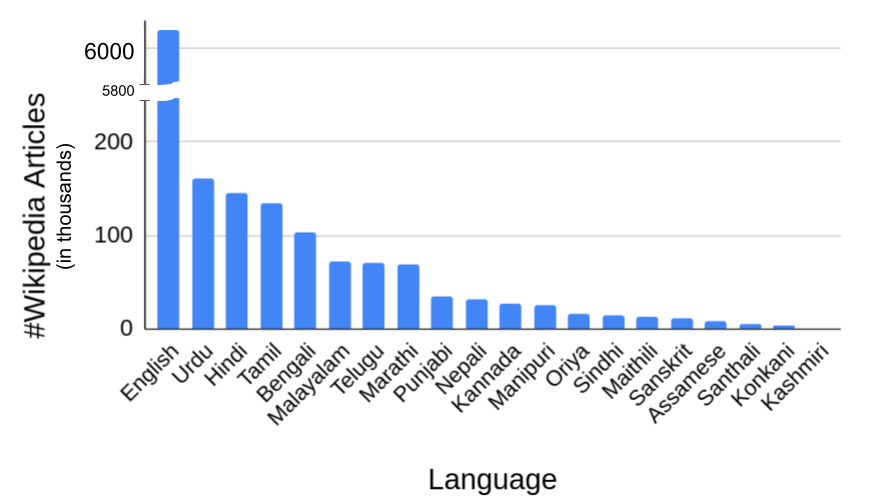}
\caption{Number of wikipedia articles for top-few Indian Languages and English. The height of the English bar is not to scale as indicated by the break. Number of English articles is roughly 400x more than articles in Oriya and 800x more than articles in Assamese.}
\label{fig:corpora-stats}
\end{figure}

\begin{figure*}[!t]
    \begin{center}
    \includegraphics[scale=0.65]{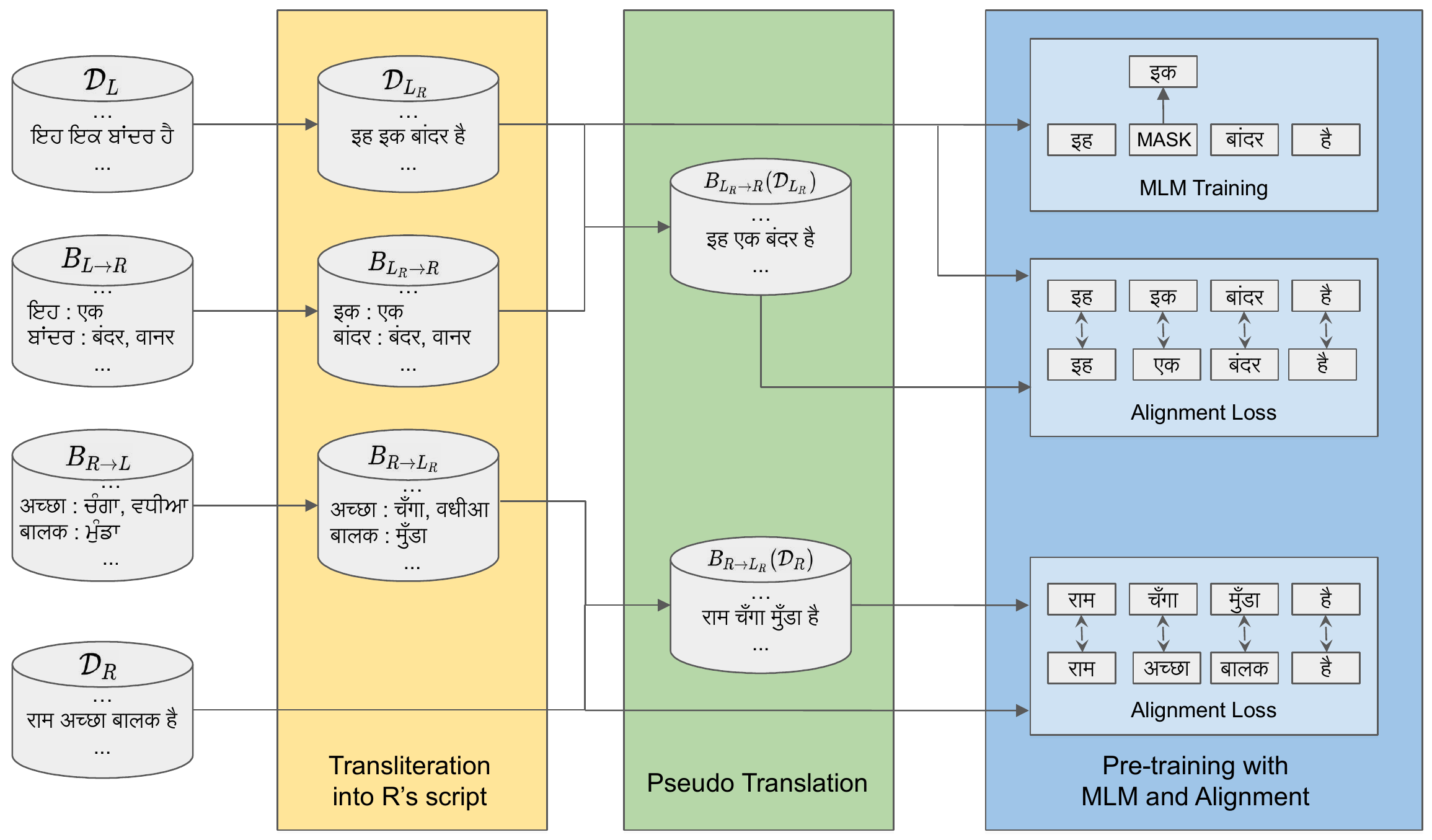}
\caption{\label{fig:overview}Pre-training with MLM and Alignment loss in \our{} with LRL $L$ as Punjabi (pa) in Gurumukhi script and \rdl\ $R$ as Hindi (hi) in Devanagari script. \our{} first transliterates LRL text in the monolingual corpus ($\dL{}$) and bilingual dictionaries ($\dLH{}$ and $\dHL{}$) to the script of the RPL $R$. The transliterated bilingual dictionaries are then used to pseudo translate the RPL corpus ($\dH{}$) and transliterated LRL corpus ($\tdL{}$). This pseudo translated data is then used to adapt the given LM $\cM{}$ for the target LRL $L$ using a combination of Masked Language Model (MLM) and alignment losses. For notations and further details, please see \refsec{sec:method}.}
    \end{center}
\end{figure*}

BERT-based pre-trained language models (LMs) have enabled significant advances in NLP \cite{devlin-etal-2019-bert,liu2019roberta,lan2020albert}. 
Pre-trained LMs have also been developed for the multilingual setting, where a single multilingual model is capable of handling inputs from many different languages. For example, the Multilingual BERT ({\mbert}) \cite{devlin-etal-2019-bert} model was trained on 104 different languages. When fine-tuned for various downstream tasks, multilingual LMs have demonstrated significant success in generalizing \emph{across} languages \cite{hu2020xtreme,conneau2019unsupervised}. Thus, such models make it possible to transfer knowledge and resources from resource rich languages to Low Web-Resource Languages (LRL). This has opened up a new opportunity towards rapid development of language technologies for LRLs.

However, there is a challenge. The current paradigm for training Mutlilingual LM requires  text corpora in the languages of interest, usually in large volumes. However, such text corpora is often available in limited quantities for LRLs. For example, in \reffig{fig:corpora-stats} we present the size of Wikipedia, a common source of corpora for training LMs, for top-few scheduled Indian languages\footnote{According to Indian Census 2011, more than 19,500 languages or dialects are spoken across the country, with 121 of them being spoken by more than 10 thousand people.} and English. The top-2 languages are just one-fiftieth the size of English,  and yet Hindi is seven  times larger than the O(20,000) documents of languages like Oriya and Assamese which are spoken by millions of people.  This calls for the development of additional mechanisms for training multilingual LMs which are not exclusively reliant on large monolingual corpora.

Recent methods of adapting a pre-trained multilingual LM to a LRL include fine-tuning the full model with an extended vocabulary~\cite{wang-etal-2020-extending}, training a light-weight adapter layer while keeping the full model fixed~\cite{pfeiffer-etal-2020-mad}, and exploiting overlapping tokens to learn embeddings of the LRL~\cite{Pfeiffer2020-unks}. These are general-purpose methods that do not sufficiently exploit the specific relatedness of languages within the same family.

We propose \textbf{\our{}} for this task. \our{} exploits \emph{relatedness} between the LRL of interest and a \textbf{\rdlfull{}} (\textbf{\rdl{}}). We focus on Indic languages, and consider Hindi as the \rdl{}.  
The languages we consider in this paper are related along several dimensions of linguistic typology~\cite{wals,littell2017}: phonologically, phylogenetically as they are all part of the Indo-Aryan family, geographically, and syntactically matching on key features like the Subject-Object-Verb (SOV) order as against the Subject-Verb-Object (SVO) order in English.  Even though the scripts of several Indic languages differ, they are all part of the same Brahmic family, making it easier to design rule-based transliteration libraries across any language pair.  In contrast,  transliteration of Indic languages to English is 
harder with considerable phonetic variation in how words are transcribed. The geographical and phylogenetic proximity has lead to significant overlap of words across languages.  This implies that just after transliteration we are able to exploit overlap with a \rdlfull{} (\rdl{}) like Hindi. On three Indic languages we discover between 11\% and 26\% overlapping tokens with Hindi, whereas with English it is less than 8\%, mostly comprising numbers and entity names.
Furthermore, the syntax-level similarity between languages allows us to generate high quality data augmentation by exploiting pre-existing bilingual dictionaries. We generate pseudo parallel data by converting \rdl\ text to LRL and vice-versa. These allow us to further align the learned embeddings across the two languages using the recently proposed 
loss functions for aligning contextual embeddings of word translations~\cite{DBLP:conf/iclr/CaoKK20,wu-dredze-2020-explicit}.



\noindent
In this paper, we make the following contributions:
\begin{itemize}
    \item We address the problem of adding a Low Web-Resource Language (LRL) to an existing pre-trained LM, especially when monolingual corpora in the LRL is limited. This is an important but underexplored problem. We focus on Indian languages which have hundred of millions of speakers, but traditionally understudied in the NLP community.
    \item We propose \our{} which exploits relatedness among languages to effectively incorporate a LRL into a pre-trained LM. We highlight the relevance of transliteration and pseudo translation for related languages, and use them effectively in \our{} to adapt a pre-trained LM to a new LRL.
    \item Through extensive experiments, we find that \our{} is able to gain significant improvements on benchmark datasets. We demonstrate how \our{} adapts {\mbert} 
    to Oriya and Assamese, two low web-resource Indian languages by pivoting through Hindi.  Via ablation studies on bilingual models we show  that \our\ is able to achieve accuracy of zero-shot transfer with limited data (20K documents) that is not surpassed even with four times as much data in existing methods.
\end{itemize}

The source code for our experiments is available at  \href{https://github.com/yashkhem1/RelateLM}{https://github.com/yashkhem1/RelateLM}.

\section{Related Work}
\label{sec:related}



Transformer \cite{vaswani2017attention} based language models like {\mbert} \cite{devlin-etal-2019-bert}, MuRIL \cite{khanuja2021muril}, IndicBERT \cite{kakwani2020inlpsuite}, and XLM-R \cite{conneau2019unsupervised}, trained on massive multilingual datasets have been shown to scale across a variety of  tasks and languages. The zero-shot cross-lingual transferability offered by these models makes them promising for low-resource domains. 
\citet{pires-etal-2019-multilingual} find that  cross-lingual transfer is even possible across languages of different scripts, but is more effective for typologically related languages. However, recent works \cite{lauscher-etal-2020-zero, pfeiffer-etal-2020-mad, hu2020xtreme} have identified poor cross-lingual transfer to languages with limited data when jointly pre-trained. A primary reason behind poor transfer is the lack of model's capacity to accommodate all languages simultaneously. This has led to increased interest in adapting multilingual LMs to  LRLs and we discuss these in the following two settings. 

\paragraph{LRL adaptation using monolingual data}
For eleven languages outside {\mbert}, ~\citet{wang-etal-2020-extending} demonstrate that adding a new target language to {\mbert} by simply extending the embedding layer with new weights results in better performing models when compared to  bilingual-BERT pre-training with English as the second language. ~\citet{Pfeiffer2020-unks} adapt multilingual LMs to the LRLs and languages with scripts unseen during pre-training by learning new tokenizers for the unseen script and initializing their embedding matrix by leveraging the lexical overlap w.r.t. the languages seen during pre-training. 
Adapter \cite{pfeiffer-etal-2020-adapterhub} based frameworks like \cite{pfeiffer-etal-2020-mad,artetxe-etal-2020-cross,ustun-etal-2020-udapter} address the lack of model's capacity to accommodate multiple languages and establish the advantages of adding language-specific adapter modules in the BERT model for accommodating LRLs. These methods generally assume access to a fair amount of monolingual LRL data and do not exploit relatedness across languages explicitly. These methods provide complimentary gains to our method of directly exploiting language relatedness. 

\paragraph{LRL adaptation by utilizing parallel data} When a parallel corpus of a high resource language and its translation into a LRL is available, ~\citet{NEURIPS2019_c04c19c2} show that pre-training on concatenated parallel sentences results in improved cross-lingual transfer. Methods like \citet{DBLP:conf/iclr/CaoKK20,wu-dredze-2020-explicit} discuss advantages of explicitly bringing together the contextual embeddings of aligned words in a translated pair. Language relatedness has been exploited in multilingual-NMT systems in various ways ~\cite{neubig-hu-2018-rapid,goyal-durrett-2019-embedding,song-etal-2020-pre}. These methods typically involve data augmentation for a LRL with help of a related high resource language ({\rdl}) or to first learn the NMT model for a {\rdl} followed by finetuning on the LRL.   
\citet{wang2019multilingual} propose a soft-decoupled encoding approach for exploiting subword overlap between LRLs and HRLs  to improve encoder representations for LRLs. \citet{gao-etal-2020-improving} address the issue of generating fluent LRL sentences in NMT by extending the soft-decoupled encoding approach to improve decoder representations for LRLs.  \citet{xia-etal-2019-generalized} utilize data augmentation techniques for LRL-English
translation using RPL-English and RPL-LRL parallel corpora induced via bilingual lexicons and unsupervised NMT. ~\citet{goyal-etal-2020-efficient} utilize transliteration and parallel data from related Indo-Aryan languages to improve NMT systems. Similar to our approach they transliterate all the Indian languages to the Devanagri script. Similarly, ~\citet{song-etal-2020-pre} utilize Chinese-English parallel corpus and transliteration of Chinese to Japanese for improving Japanese-English NMT systems via data augmentation.  \\
\noindent
To the best of our knowledge no earlier work has explored the surprising effectiveness of transliteration to a related existing prominent 
language, for learning multilingual LMs, although some work exists in NMT as mentioned above.

\section{Low Web-Resource Adaptation in \our}
\label{sec:method}


\begin{table}[t]
        \centering
        \begin{adjustbox}{width=0.45\textwidth}
        \begin{tabular}{|c||c|c|}
        \hline 
         \multicolumn{3}{|c|}{Percentage Overlap of Words} \\
         \hline
        LRL & Related Prominent & Distant Prominent  \\
         &  (Hindi) &  (English) \\
        \hline 
        Punjabi  & \textbf{25.5} &  7.5\\
        Gujarati & \textbf{23.3} &  4.5\\
        Bengali  & \textbf{10.9} &  5.5\\      
        \hline 
       \end{tabular}
      \end{adjustbox}
       \caption{\label{wordoverlap} \textbf{Motivation for transliteration}: 
       \% overlapping words between transliterated LRL (in Prominent Language's script) and prominent language text. \% overlap is defined as the number of common distinct words divided by number of distinct words in the transliterated LRL. Overlap is much higher with Hindi, the Related Prominent Language (\rdl{}), compared to English, the distant language. Overlapping words act as anchors during multilingual pre-training in \our{}(\refsec{sec:transliteration})}
\end{table}
\begin{table}[t]
        \centering
        \begin{adjustbox}{width=0.45\textwidth}
        \begin{tabular}{|c||c|c|}
        \hline
        \multicolumn{3}{|c|}{BLEU Scores} \\
         \hline
        LRL & Related Prominent & Distant Prominent  \\
        (Target) &  (Hindi) (Source) &  (English) (Source) \\
        \hline
        Punjabi  & \textbf{24.6}                         & 16.5                           \\
        Gujarati & \textbf{20.3}                         & 12.9                           \\
        Bengali  & \textbf{19.3}                         & 12.4                          \\
        \hline 
       \end{tabular}
      \end{adjustbox}
       \caption{\label{bleu}\textbf{Motivation for pseudo translation}: 
       BLEU scores between pseudo translated prominent language sentences and LRL sentences. BLEU with  Hindi, the \rdl{}, is much higher than with English, the distant prominent language highlighting the effectiveness of pseudo translation from a \rdl{}
%
(\refsec{sec:pseudo}). English and Hindi dictionary sizes same. For these experiments, we used a parallel corpus across these 5 languages obtained from TDIL (\refsec{sec:setup})}
\end{table}

\paragraph{Problem Statement and Notations}
Our goal is to augment an existing multilingual language model $\cM$, for example {\mbert}, to learn representations for a new LRL $\cL$ for which available monolingual corpus $\dL$ is limited.   We are also told that the language to be added is related to another language $\cH$  on which the model  $\cM$ is already  pre-trained, and is of comparatively higher resource.   However, the script of $\dL$ may be distinct from the scripts of existing languages in $\cM$.  In this section we present strategies for using this knowledge to better adapt $\cM$ to $\cL$ than the existing baseline of
fine-tuning $\cM$ using the
standard masked language model (MLM) loss on the limited monolingual data $\dL$~\cite{wang-etal-2020-extending}.   In addition to the monolingual data $\dH$ in the \rdl{} and $\dL$ in the LRL, we have access to a limited bilingual lexicon $\dLH$ that map a word in language $\cL$ to a list of synonyms in language $\cH$  and vice-versa $\dHL$.

We focus on the case where the \rdl, LRL  pairs are part of the Indo-Aryan language families where several levels of relatedness exist.  Our proposed approach, consists of three steps, viz., Transliteration to \rdl{}'s script, Pseudo translation, and Adaptation through Pre-training.  We describe each of these steps below.  \reffig{fig:overview} presents an overview of our approach.



\subsection{Transliteration}
\label{sec:transliteration}

First, the scripts of Indo-Aryan languages are part of the same Brahmic script. 
This makes it easier to design simple rule-based transliterators to convert a corpus in one script to another.  For most languages transliterations are easily available. 
Example, the Indic-Trans Library~\footnote{\url{https://github.com/libindic/indic-trans}}~\cite{Bhat:2014:ISS:2824864.2824872}. We use $\tdL$ to denote the LRL corpus after transliterating to the script of the \rdl{}. 
We then propose to further pre-train the model $\cM$ with MLM on the transliterated corpus $\tdL$ instead of $\dL$. Such a strategy could provide little additional gains over the baseline, or could even hurt accuracy, if the two languages were not sufficiently related.  For languages in the Indo-Aryan family because of strong phylogenetic and geographical overlap, many words across the two languages overlap and preserve the same meaning.  
In Table~\ref{wordoverlap} we provide statistics of the overlap of words across several transliterated Indic languages with Hindi and English.  Note that for Hindi the fraction of overlapping words is much higher than with English which are mostly numbers, and entity names. 
These overlapping words serve as anchors to align the representations for the non-overlapping words of the LRL that share semantic space with words in the \rdl.

\subsection{Pseudo Translation with Lexicons}
\label{sec:pseudo}

Parallel data between a \rdl{} and LRL language pair has been shown to be greatly useful for efficient adaptation to LRL~\cite{NEURIPS2019_c04c19c2,DBLP:conf/iclr/CaoKK20}.  However, creation of parallel data requires expensive supervision, and is not easily available for many low web-resource languages.  Back-translation is a standard method of creating pseudo parallel data but for low web-resource languages we cannot assume the presence of a well-trained translation system.  We exploit the relatedness of the Indic languages to design a pseudo translation system that is motivated by two factors:

\begin{itemize}
    \item 
    First, for most geographically proximal \rdl\-LRL language pairs, word-level bilingual dictionaries have traditionally been available to enable communication.  When they are not, crowd-sourcing creation of word-level dictionaries\footnote{Wiktionary is one such effort} requires lower skill and resources than sentence level parallel data.  Also, word-level lexicons can be created semi-automatically~\cite{zhang-etal-2017-adversarial}~\cite{artetxe-etal-2019-bilingual}~\cite{xu-etal-2018-unsupervised}.
    \item Second, Indic languages exhibit common syntactic properties that control how words are composed to form a sentence.  For example, they usually follow the Subject-Object-Verb (SOV) order as against the Subject-Verb-Object (SVO) order in English.
\end{itemize}   

We therefore create pseudo parallel data between $\cH$ and $\cL$ via a simple word-by-word translation using the bilingual lexicon. In a lexicon a word can be mapped to multiple words in another language.  We choose a word with probability proportional to its frequency in the  monolingual corpus $\dL$.  We experimented with a few other methods of selecting words that we discuss in Section~\ref{sec:expt:lookup}. 
In Table~\ref{bleu} we present BLEU scores obtained by our pseudo translation model of three Indic languages from Hindi and from English.  We observe much high BLEU for translation from Hindi highlighting the syntactic relatedness of the languages.

Let $(\dH, \tdHL(\dH))$ denote the parallel corpus formed by pseudo translating the \rdl{} corpus via the transliterated \rdl{} to LRL lexicon.  Likewise let $(\tdL, \tdLH(\tdL))$ be formed by pseudo translating the transliterated low web-resource corpus via the transliterated LRL to \rdl{} lexicon.
\subsection{Alignment Loss}
\label{sec:model-training}

The union of the two pseudo parallel corpora above, collectively called $\cP$, is used  for fine-tuning $\cM$ using an alignment loss similar to the one proposed in \cite{DBLP:conf/iclr/CaoKK20}.  This loss 
attempts to bring the multilingual embeddings of different languages closer by aligning
the corresponding word embeddings of the source language sentence and the pseudo translated target language sentence.  Let $\mathcal{C}$ be a random batch of source and (pseudo translated) target sentence pairs from $\cP$, i.e. ${\mathcal{C}}$ = $(({\bf s}^1,{\bf t}^1),({\bf s}^2,{\bf t}^2),...,({\bf s}^N,{\bf t}^N))$,
where ${\bf s}$ and ${\bf t}$ are the source and target sentences respectively. 
Since our parallel sentences are obtained via word-level translations, the alignment among words is known and monotonic.  Alignment loss has two terms:

     $\mathcal{L} = \mathcal{L}_{align} + \mathcal{L}_{reg}$
where $\mathcal{L}_{align}$ is used to bring the contextual embeddings closer and $\mathcal{L}_{reg}$ is the regularization loss which
prevents the new embeddings from deviating far away from the pre-trained embeddings. Each of these are defined below: 
\begin{equation*}
    \mathcal{L}_{align} = \sum_{({\bf s},{\bf t})\in\mathcal{C}} \sum_{i=1}^{\text{\#word}({\bf s})}||f({\bf s},l_{{\bf s}}(i)) - f({\bf t},l_{{\bf t}}(i))||_2^2 
\end{equation*}
\vspace{-8pt}
\begin{equation*}
\begin{aligned}[b]
    \mathcal{L}_{reg} = & \sum_{({\bf s},{\bf t})\in\mathcal{C}} \left(\sum_{j=1}^{\text{\#tok}({\bf s})} ||(f({\bf s},j)-f_{0}({\bf s},j)||_2^2\right.\\
    &+\left.\sum_{j=1}^{\text{\#tok}({\bf t})}||f({\bf t},j)-f_{0}({\bf t},j)||_2^2\right)
\end{aligned}
\end{equation*}
where $l_{{\bf s}}(i)$ is the position of the last token of i-th word in sentence ${\bf s}$ and $f({\bf s},j)$ is the learned contextual embedding of token at $j$-th position in sentence ${\bf s}$, i.e, for $\mathcal{L}_{align}$ we consider only the last tokens of words in a sentence, while for $\mathcal{L}_{reg}$ we consider all the tokens in the sentence.
$f_{0}({\bf s},j)$ denotes the fixed pre-trained contextual embedding of the token at $j$-th position in sentence ${\bf s}$. $\text{\#word}({\bf s})$ and $\text{\#tok}({\bf s})$ are the number of (whole) words and tokens in sentence ${\bf s}$ respectively. 



\section{Experiments}
\label{sec:expts}

We carry out the following experiments to evaluate \our{}'s effectiveness in LRL adaptation:

\begin{itemize}
    \item First, in the full multilingual setting, we evaluate whether \our{} is capable of extending mBERT with two unseen low-resource Indic languages: Oriya (unseen script) and Assamese (seen script).  (\refsec{sec:expt-multilm})
    \item We then move to the bilingual setting where we use \our{} to adapt a model trained on a single \rdl{} to a LRL. This setting allowed us to cleanly study the impact of different adaptation strategies and experiment with many \rdl-LRL language pairs. (\refsec{sec:expt-bilm})
    \item Finally, \refsec{sec:expt:lookup}, presents an ablation study on dictionary lookup methods, alignment losses, and corpus size.
\end{itemize}
   We evaluate by measuring the efficacy of zero-shot transfer from the \rdl\ on three different tasks: NER, POS and text classification.

\subsection{Setup}
\label{sec:setup}
\paragraph{LM Models}
We take m-BERT as the model $\cM$ for our multilingual experiments. For the bilingual experiments, we start with two separate monolingual language models on  each of Hindi and English language to serve as $\cM$. For Hindi  we trained our own \hbert\ model over the 160K monolingual Hindi Wikipedia articles using a vocab size of 20000 generated using WordPiece tokenizer.  For English we use the pre-trained BERT model which is trained on almost two orders of magnitude Wikipedia articles and more. When the LRL is added in its own script, we use the bert-base-cased model and when the LRL is added after transliteration to English, we use the bert-base-uncased model.

\paragraph{LRLs, Monolingual Corpus, Lexicon}
As LRLs we consider five Indic languages spanning four different scripts.  Monolingual data was obtained from Wikipedia as summarized in Table~\ref{tab:script}.
We extend m-BERT with two unseen low web-resource languages: Assamese and Oriya.  Since it was challenging to find Indic languages with task-specific labeled data but not already in m-BERT, we could not evaluate on more than two languages.  For the bilingual model experiments, we adapt each of \hbert\ and English BERT with three different languages: Punjabi, Gujarati and Bengali.  For these languages we simulated the LRL setting by 
downsampling their Wikipedia data to 20K documents. For experiments where we require English monolingual data for creating pseudo translations, we use a downsampled version of English Wikipedia having the same number of documents as the Hindi Wikipedia dump.

The addition of a new language to $\cM$ was done by adding 10000 tokens of the new language generated by WordPiece tokenization to the existing vocabulary, with random initialization of the new parameters. For all the experiments, we use libindic’s indictrans library~\cite{Bhat:2014:ISS:2824864.2824872} for transliteration. For pseudo translation we use the union of Bilingual Lexicons obtained from CFILT \footnote{\url{https://www.cfilt.iitb.ac.in/}} and Wiktionary \footnote{\url{https://hi.wiktionary.org/wiki/}} and their respective sizes for each language are summarized in Table~\ref{tab:script}

\begin{table}[t]
    \centering
    \begin{adjustbox}{width=0.48\textwidth}
    \begin{tabular}{|l|l|c|c|c|}
    \hline
        
        \multirow{2}{*}{Dataset Split} & \multirow{2}{*}{Lang} & \multicolumn{3}{|c|}{Number of Sentences}  \\ \cline{3-5}
        & & NER & POS & TextC. \\  \hline
        \multirow{2}{*}{Train Data \rdl} 
                        & en & 20.0 & 56.0 & 27.0 \\
                        & hi & 5.0 & 53.0 & 25.0 \\ \hline
        \multirow{2}{*}{Val Data \rdl} & en & 10.0 & 14.0 & 3.8 \\ 
         & hi & 1.0 & 13.0 & 4.0 \\ \hline
        \multirow{5}{*}{Test Data LRL} & pa & 0.2 & 13.4 & 7.9 \\
         & gu & 0.3 & 14.0 & 8.0 \\ 
         & bn & 1.0 & 9.7 & 5.8 \\
         & as & - & 14.0 & 8.0 \\ 
         & or & 0.2 & 4.0 & 7.6 \\ \hline
    \end{tabular}
    \end{adjustbox}
     \caption{Statistics of Task-specific Datasets. All numbers are in thousands.}
       \label{tab:data}
\end{table}

\begin{table}[!t]
        \centering
        \begin{adjustbox}{width=0.48\textwidth}
        \begin{tabular}{|l|r|c|r|r|r|r|}
        \hline
\multirow{2}{*}{LRL} & \multirow{2}{*}{\#Docs} & \multirow{2}{*}{Scripts} & \multicolumn{2}{c|}{hi-Lexicon} & \multicolumn{2}{c|}{en-Lexicon} \\ \cline{4-7} 
                      &                          &                          & Fw              & Bw             & Fw              & Bw              \\ \hline
pa                    & 20                       & Gurumukhi                & 53             & 65             & 18             & 15             \\ 
gu                    & 20                       & Gujarati                 & 29             & 43             & 18             & 10             \\ 
bn                    & 20                       & As-Bangla                   & 23             & 40             & 12             & 10             \\ 
or                    & 20                       & Oriya                     & 18             & 18             & 18             & 18             \\ 
as                    & 7                      & As-Bangla                   & 19             & 17             & 19             & 17             \\ \hline

       \end{tabular}
      \end{adjustbox}
      \caption{Statistics of resources used for LRLs in the experiments. All the numbers are in thousands.  \#Docs represents number of documents for each language. For each language, hi-Lexicon and en-Lexicon report sizes of bilingual Hindi and English dictionaries respectively in either direction. Fw represents the direction from a LRL to hi or en. Hindi uses the Devanagri script with a vocab size of 20K. For all other languages the vocab size is fixed at 10K. As-Bangla refers to the Bengali-Assamese script.}
    \label{tab:script}
\end{table}

\begin{table*}[t]
        \centering
        \begin{adjustbox}{width=0.95\textwidth}
        \begin{tabular}{|l|c|l|l|l|l|l|l|l|l|l|}
        \hline
        \multirow{2}{*}{LRL Adaptation} & \multirow{2}{*}{\makecell{Prominent\\Language}} & \multicolumn{3}{|c|}{\bf Punjabi}& \multicolumn{3}{|c|}{\bf Gujarati}& \multicolumn{3}{|c|}{\bf Bengali}\\
        \cline{3-11}
         & & NER & POS & TextC. & NER & POS & TextC. & NER & POS & TextC. \\
        \hline 
        
        \gray{mBERT} & \gray{-} & \gray{41.7} & \gray{86.3} & \gray{64.2}  & \gray{39.8} & \gray{87.8} & \gray{65.8} & \gray{70.8} & \gray{83.4} & \gray{75.9}\\  
         \hline
         \hline
        EBERT {\scriptsize \cite{wang-etal-2020-extending}} & \multirow{2}{*}{en} & 19.4 & 48.6 & 33.6 & 14.5  & 56.6 & 37.8 & 31.2  & 50.7 &  32.7\\

        \our$-$PseudoT &  & 38.6  & 58.1 & 54.7 & 15.3 & 58.5 & 57.2 & \textbf{68.8}  & 59.8 & 58.6\\
        \hline
        EBERT {\scriptsize \cite{wang-etal-2020-extending}} & \multirow{3}{*}{hi} & 28.2 & 78.6 & 51.4 & 14.8 & 69.0  &  48.1 & 34.0 & \textbf{73.2}  &  45.6 \\
        \our$-$PseudoT &  & 65.1 & 77.3  & 76.1  & 39.6 & 80.2 & 79.1 & 56.3 & 69.9  & 77.5 \\
        \our &   & \textbf{66.9} & \textbf{81.3} & \textbf{78.6} & \textbf{39.7} & \textbf{82.3} & \textbf{79.8} & 57.3 & 71.7 & \textbf{78.7}\\

        \hline
        \end{tabular}
      \end{adjustbox}
       \caption{Different Adaptation Strategies evaluated for zero-shot transfer (F1-score) on NER, POS tagging and Text Classification after fine-tuning with the Prominent Language (English or Hindi). mBERT, which is trained with much larger datasets and more languages is not directly comparable, and is presented here  just for reference.}
       \label{overall}
\end{table*}

\paragraph{Tasks for zero-shot transfer evaluation}
After adding a LRL in $\cM$, we perform 
task-specific fine-tuning on the \rdl\ separately for three tasks: NER, POS and Text classification.  Table~\ref{tab:data} presents a summary of the training, validation data in \rdl\ and test data in LRL on which we perform zero-shot evaluation.
We obtained the NER data from WikiANN \cite{Pan2017} and XTREME \cite{hu2020xtreme} and the POS and Text Classification data from 
the Technology Development for Indian Languages (TDIL)\footnote{\url{https://www.tdil-dc.in}}. We downsampled the TDIL data for each language to make them class-balanced. The POS tagset used was the BIS Tagset \cite{sardesai-etal-2012-bis}. For the English POS Dataset, we had to map the PENN tagset in  to the BIS tagset. We have provided the mapping that we used in the Appendix (\hyperref[sec:tagsetmapping]{B})

\paragraph{Methods compared} 
We contrast \our\ with three other adaptation techniques: (1) EBERT~\cite{wang-etal-2020-extending} that extends the vocabulary and tunes with MLM on $\dL$ as-is, (2) \our\ without pseudo translation loss, and (3) m-BERT when the language exists in m-BERT.

\paragraph{Training Details} For pre-training on MLM we chose batch size as 2048, learning rate as 3e-5 and maximum sequence length as 128. We used whole word masking for MLM and BertWordPieceTokenizer for tokenization. For pre-training Hi-BERT the duplication was taken as 5 with training done for 40K iterations. For all LRLs where monolingual data used was 20K documents, the  duplication factor was kept at 20 and and training was done for 24K iterations. For Assamese, where monolingual data was just 6.5K documents, a duplication factor of 60 was used with the same 24K training iterations. The MLM pre-training was done on Google v3-8 Cloud TPUs.

For alignment loss on pseudo translation we chose learning-rate as 5e-5, batch size as 64 and maximum sequence length as 128. The training was done for 10 epochs also on Google v3-8 Cloud TPUs.
For task-specific fine-tuning we used learning-rate 2e-5 and batch size 32, with training duration as 10 epochs for NER, 5 epochs for POS and 2400 iterations for Text Classification. The models were evaluated on a separate RPL validation dataset and the model with the minimum F1-score, accuracy and validation loss was selected for final evaluation for NER, POS and Text Classification respectively. All the fine-tuning experiments were done on Google Colaboratory. The results reported for all the experiments are an average of 3 independent runs.

\begin{table}[t]
        \centering
        \begin{adjustbox}{width=0.48\textwidth}
        \begin{tabular}{|l||c|c|c|c|}
        \hline
        LRL adaptation & Prominent & NER & POS & TextC. \\
                      & Language    &     &      &    \\
                      \hline 
        \multicolumn{5}{|l|}{\bf Oriya}\\
        \hline
        \our$-$PseudoT & \multirow{2}{*}{en} & 14.2 & 72.1 & 63.2 \\
        \our &  & 16.4 & 74.1 & 62.7\\
        \hline
        EBERT {\scriptsize \cite{wang-etal-2020-extending}} & \multirow{3}{*}{hi} & 10.8 & 71.7 & 53.1\\
        \our$-$PseudoT &  & 22.7 & 74.7 & 76.5 \\
        \our &  & {\bf 24.7} & {\bf 75.2} & {\bf 76.7} \\

        \hline
        \hline
        \multicolumn{5}{|l|}{\bf Assamese}\\
        \hline
        \our$-$PseudoT & \multirow{2}{*}{en} & - & 78.2  & 74.8  \\
        \our & & - & 77.4 & 74.7 \\
        \hline
        EBERT {\scriptsize \cite{wang-etal-2020-extending}} & \multirow{3}{*}{hi} & - & 71.9 & 78.6\\
        \our$-$PseudoT &  & - & {\bf 79.4} & 79.8 \\
        \our & & -  & 79.3 & {\bf 80.2} \\
        \hline

      \end{tabular}
      \end{adjustbox}
      \caption{mBERT+LRL with different adaptation strategies evaluated on NER, POS tagging and Text Classification with both English and Hindi as the fine-tuning languages.  Accuracy metric is F1.}
      \label{tab:mbert}
\end{table}

\subsection{Multilingual Language Models}
\label{sec:expt-multilm}

\begin{figure*}[t]
    \centering
  \begin{subfigure}[b]{0.32\textwidth}
    \includegraphics[width=\textwidth]{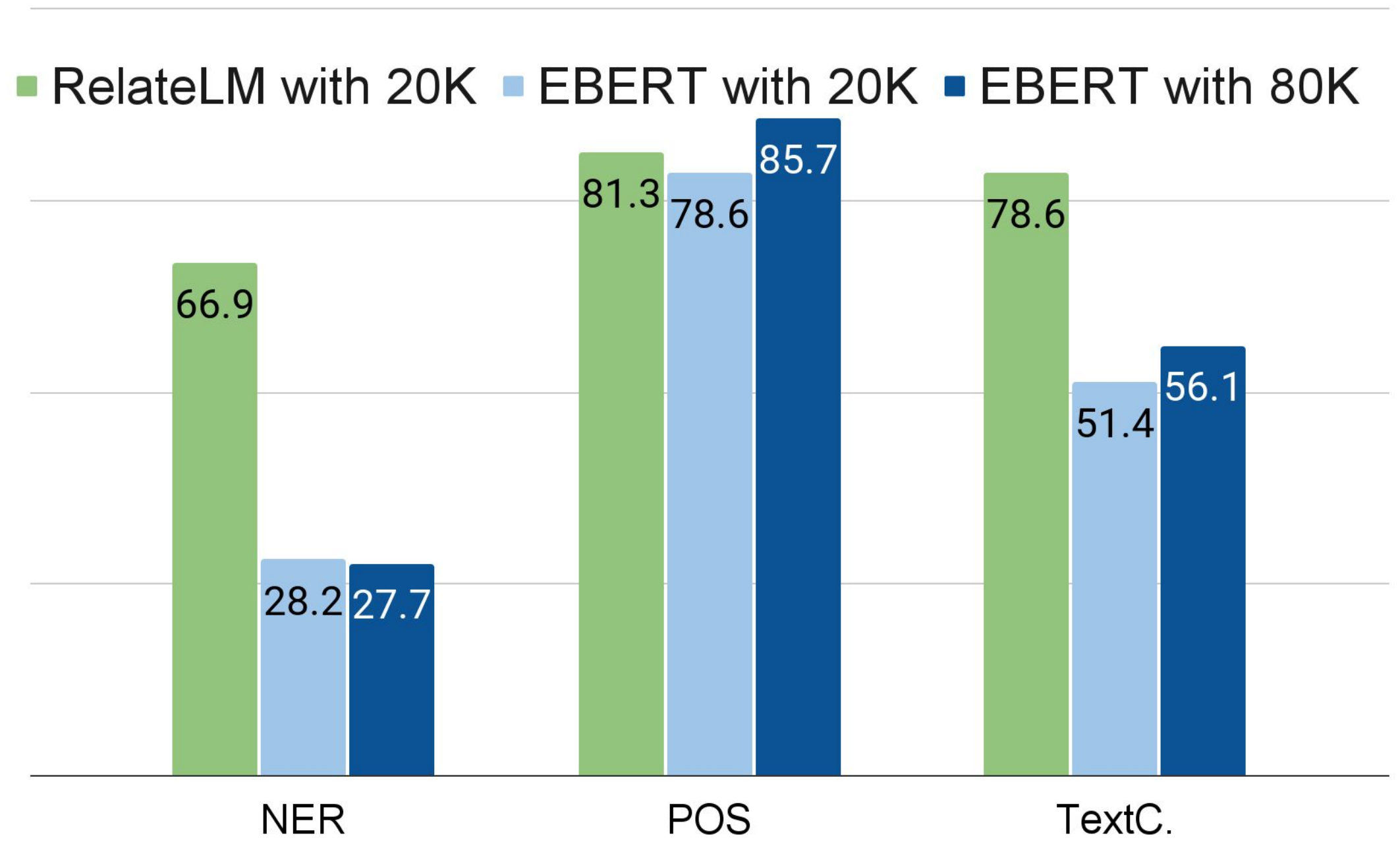}
    \caption{Punjabi}
    \label{fig:f1}
  \end{subfigure}
  \begin{subfigure}[b]{0.32\textwidth}
    \includegraphics[width=\textwidth]{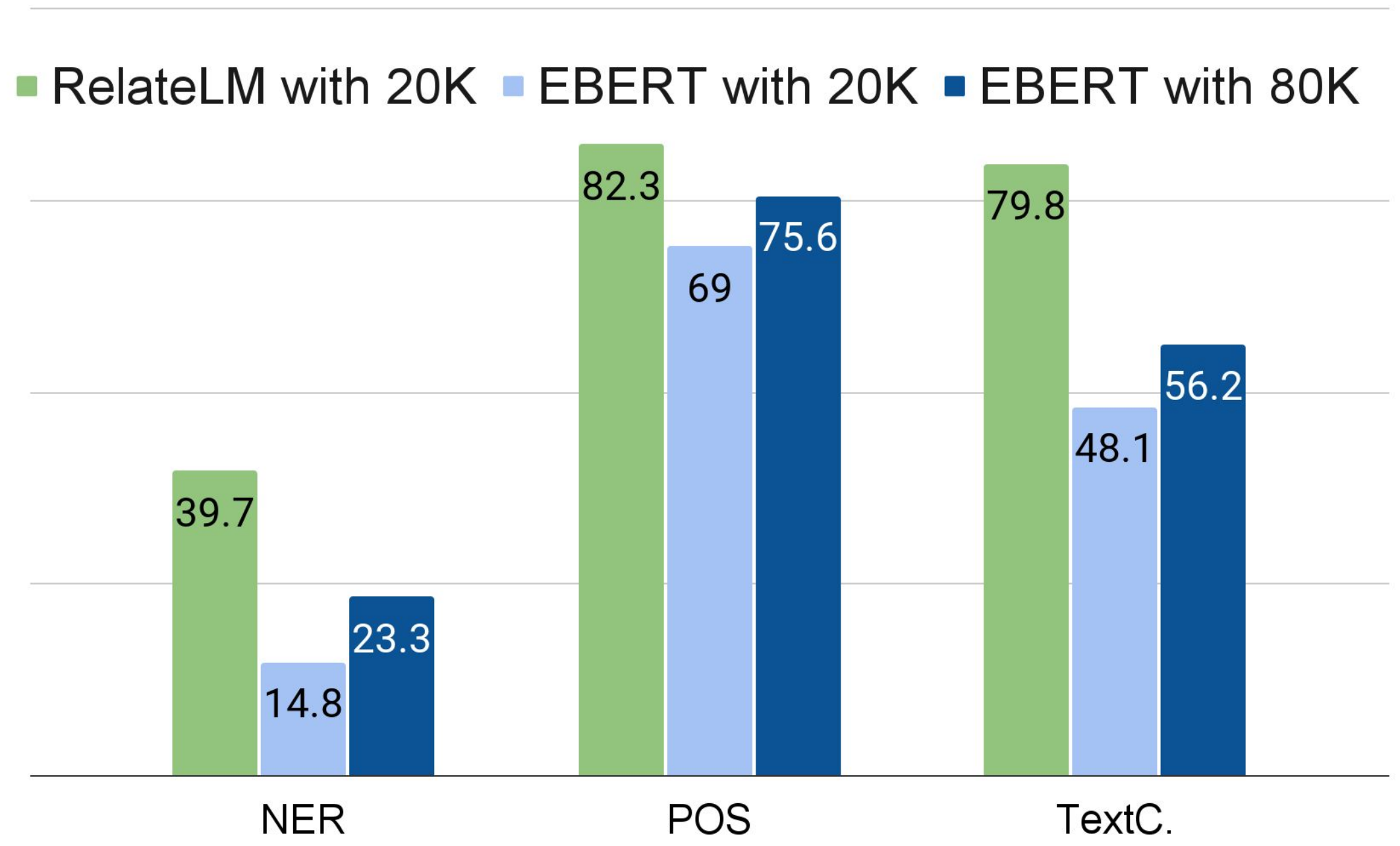}
    \caption{Gujarati}
    \label{fig:f2}
  \end{subfigure}
  \begin{subfigure}[b]{0.32\textwidth}
    \includegraphics[width=\textwidth]{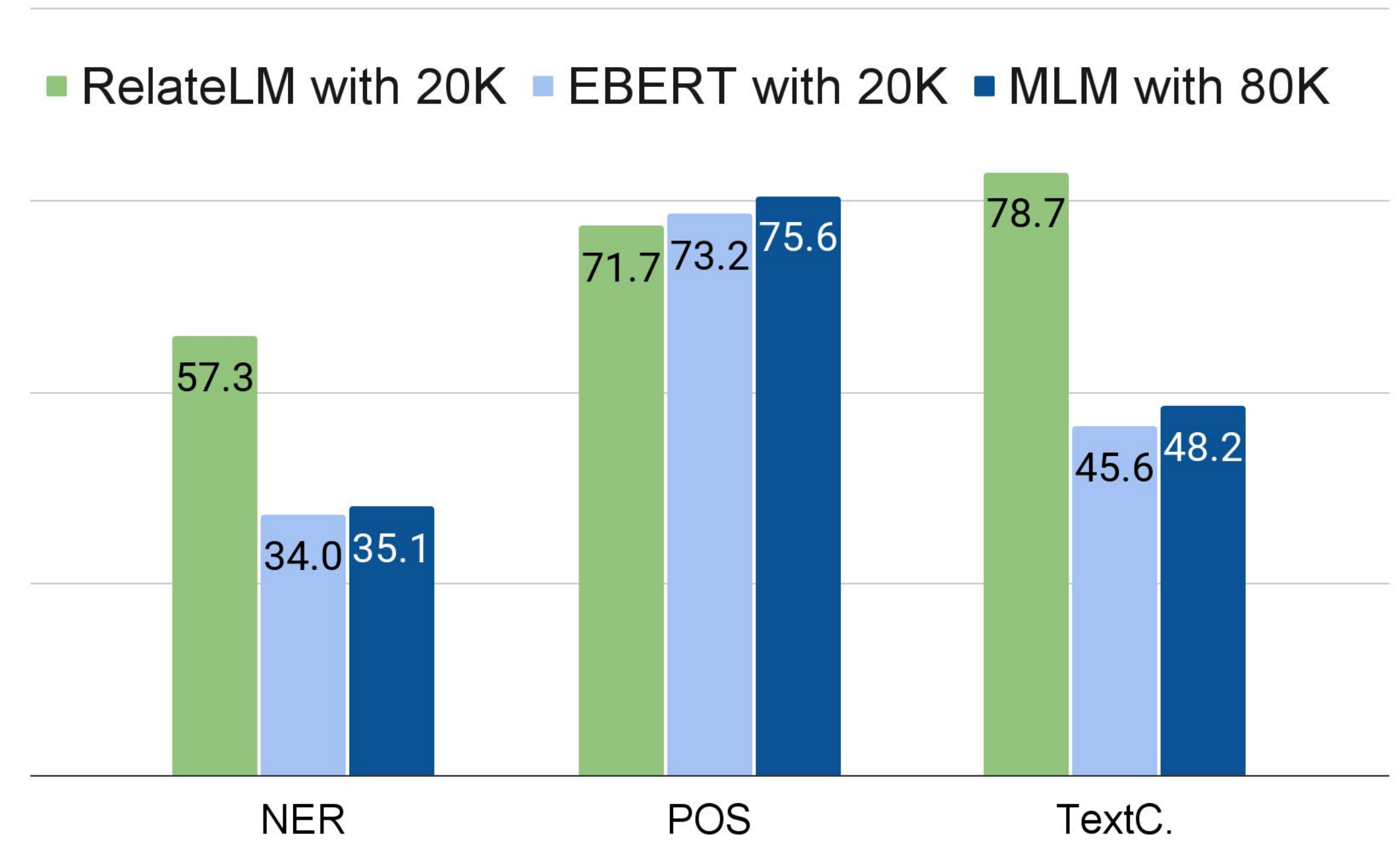}
    \caption{Bengali}
    \label{fig:f3}
  \end{subfigure}
  \caption{Comparison of F1-score between \our{}-20K, EBERT-20K and EBERT-80K, where the number after method name indicates pre-training corpus size. We find that \our{}-20K outperforms EBERT-20K in 8 out of 9 settings, and even outperforms EBERT-80K, which is trained over 4X more data, in 7 out of 9 settings.}
  \label{fig:corpus:size}
\end{figure*}

We evaluate \our's adaptation strategy on mBERT, a state of the art multilingual model with two unseen languages: Oriya and Assamese.
%
The script of Oriya is unseen whereas the script of Assamese is the same as Bengali (already in m-BERT).   Table~\ref{tab:mbert} compares  different adaptation strategies including the option of treating each of Hindi and English as \rdl\ for transliteration into.  For both LRLs, transliterating to Hindi as \rdl\ provides gains over EBERT that keeps the script as-is and English transliteration. We find that these gains are much more significant for Oriya than Assamese, which could be because Oriya is a new script. Further augmentation with pseudo translations with Hindi as \rdl, provides significant added gains. We have not included the NER results for Assamese due to the absence of good quality evaluation dataset.

\subsection{Bilingual Language Models}
\label{sec:expt-bilm}
For more extensive experiments and ablation studies we move to bilingual models.
Table~\ref{overall} shows the results of different methods of adapting $\cM$ to a LRL with \hbert\ and \ebert\ as two choices of $\cM$. 
We obtain much higher gains when the LRL is transliterated to Hindi than to English or keeping the script as-is. This suggests that transliteration to a related language succeeds in parameter sharing between a \rdl\ and a LRL.  Note that the English BERT model is trained on a much larger English corpus than the \hbert\ model is trained on the Hindi corpus.  Yet, because of the relatedness of the languages we get much higher accuracy when adding transliterated data to Hindi rather than to English.  
Next observe that pre-training with alignment loss on pseudo translated sentence pairs improves upon the results obtained with transliteration. This shows that  pseudo translations is a decent alternative when a  parallel translation corpora is not available.

Overall, we find that \our\ provides substantial gains over the baseline.  In many cases \our\ is even better than mBERT which was pre-trained on a lot more monolingual data in that language. 
Among the three languages, we obtain lowest gains for Bengali since 
the phonetics of Bengali varies to some extent from other Indo-Aryan languages, and Bengali shows influence from Tibeto-Burman languages too~\cite{kunchukuttan2020utilizing}. This is also evident in the lower word overlap and lower BLEU in Table~\ref{wordoverlap} and Table~\ref{bleu} compared to other Indic languages. We further find that in case of Bengali, the NER results are best when Bengali is transliterated to English rather than Hindi, which we attribute to the presence of English words in the NER evaluation dataset.

\begin{table}[!t]
        \centering
        \begin{adjustbox}{width=0.40\textwidth}
        \begin{tabular}{|l|l|l|l|l|}
        \hline
        Loss & Dict Lookup & NER & POS &  Text C.\\
        \hline
        \multicolumn{5}{|l|}{\bf Punjabi}\\
        \hline
        MSE & first  & 62.4 &  80.0 & 77.6\\
        MSE & max  & 68.2 &  81.3 &  77.6\\
        MSE & root-weighted  & 64.9 &  78.9 &  76.9\\
        MSE &  weighted &  66.9  & 81.3  & 78.6\\
        cstv &  weighted &  68.2  & 80.8  & 79.4\\
        \hline
        \hline 
        \multicolumn{5}{|l|}{\bf Gujarati}\\
        \hline
        MSE & first  & 39.2 &  83.3 &  78.6\\
        MSE & max  & 39.1  & 82.5 &  80.4\\
        MSE & root-weighted & 39.7 & 82.6 &  79.9\\
        MSE & weighted & 39.7 & 82.3 &  79.8\\
        cstv & weighted & 40.2 & 84.0 &  81.6\\
        \hline
        \hline 
        \multicolumn{5}{|l|}{\bf Bengali}\\
        \hline
        MSE & first & 55.5 & 68.0 & 74.0\\
        MSE & max  & 56.2 &  70.3 & 79.7\\
        MSE & root-weighted  & 56.4 & 69.3 &  76.5\\
        MSE & weighted & 57.3 & 71.7 & 78.7\\
        cstv & weighted & 56.6 & 67.6 & 76.5\\
        \hline
      \end{tabular}
      \end{adjustbox}
      \caption{Usefulness of Bilingual Dictionaries with MSE(Mean Squared Error Loss) and cstv(Contrastive Loss) evaluated on NER, POS tagging and Text Classification in \our.} 
      \label{lookup}
\end{table}
\subsection{Ablation Study}
\label{sec:expt:lookup}

\paragraph{Methods of Dictionary Lookups}
We experimented with various methods of choosing the translated word from the lexicon which may have multiple entries for a given word. In Table~\ref{lookup} we compare four methods of picking entries: {\it first} - entry at first position, {\it max}-entry with maximum frequency in the monolingual data, {\it weighted} - entry with probability proportional to that frequency and {\it root-weighted} - entry with probability proportional to the square root of that frequency. We find that these four methods are very close to each other, with the {\it weighted} method having a slight edge.
\paragraph{Alignment Loss}
We compare the MSE-based loss we used with the recently proposed contrastive loss~\cite{wu-dredze-2020-explicit} for $\mathcal{L}_{align}$ but did not get any significant improvements. We have provided the results for additional experiments in the Appendix (\hyperref[sec:cstvlossexp]{A})
\paragraph{Increasing Monolingual size}
In Figure~\ref{fig:corpus:size} we increase the monolingual LRL data used for adapting EBERT four-fold and compare the results. We observe that even on increasing monolingual data, in most cases, by being able to exploit language relatedness, \our\ outperforms the EBERT model with four times more data. These experiments show that for zero-shot generalization on NLP tasks, it is more important to improve the alignment among languages by exploiting their relatedness, than to add more monolingual data.  




\section{Conclusion and Future Work}
\label{sec:conclusion}
We address the problem of adapting a pre-trained language model (LM) to a Low Web-Resource Language (LRL) with limited monolingual corpora. 
We propose \our{}, which explores  \emph{relatedness} between the LRL and a \rdlfull{} (\rdl{}) already present in the LM. \our{} exploits relatedness along two dimensions -- script relatedness through transliteration, and sentence structure relatedness through pseudo translation.  We focus on Indic languages, which have hundreds of millions of speakers, but are understudied in the NLP community. Our experiments provide evidence that \our{} is effective in adapting  multilingual LMs (such as mBERT) to various LRLs.  Also, \our{} is able to achieve zero-shot transfer with limited LRL data (20K documents) which is not surpassed even with 4X more data by existing baselines. Together, our experiments establish that using a related language as pivot, along with data augmentation through transliteration and bilingual dictionary-based pseudo translation, can be an effective way of adapting an LM for LRLs, and that this is more effective than direct training or pivoting through English.

Integrating \our{} with other complementary methods for adapting LMs for LRLs \cite{pfeiffer-etal-2020-mad,Pfeiffer2020-unks} is something we plan to pursue next. We are hopeful that the idea of utilizing relatedness to adapt LMs for LRLs will be effective in adapting LMs to LRLs in other languages families,  such as South-east Asian and Latin American languages. We leave that and exploring other forms of relatedness as fruitful avenues for future work.

\paragraph*{Acknowledgements}
We thank Technology Development for Indian Languages (TDIL) Programme initiated by the Ministry of Electronics \& Information Technology, Govt. of India for providing us datasets used in this study. The experiments reported in the paper were made possible by a Tensor Flow Research Cloud (TFRC) TPU grant. The IIT Bombay authors thank Google Research India for supporting this research.  We thank Dan Garrette and Slav Petrov for providing comments on an earlier draft. 

\bibliographystyle{acl_natbib}
\bibliography{paper}

\begin{thebibliography}{35}
\expandafter\ifx\csname natexlab\endcsname\relax\def\natexlab#1{#1}\fi

\bibitem[{Artetxe et~al.(2019)Artetxe, Labaka, and
  Agirre}]{artetxe-etal-2019-bilingual}
Mikel Artetxe, Gorka Labaka, and Eneko Agirre. 2019.
\newblock \href {https://doi.org/10.18653/v1/P19-1494} {Bilingual lexicon
  induction through unsupervised machine translation}.
\newblock In \emph{Proceedings of the 57th Annual Meeting of the Association
  for Computational Linguistics}, pages 5002--5007, Florence, Italy.
  Association for Computational Linguistics.

\bibitem[{Artetxe et~al.(2020)Artetxe, Ruder, and
  Yogatama}]{artetxe-etal-2020-cross}
Mikel Artetxe, Sebastian Ruder, and Dani Yogatama. 2020.
\newblock \href {https://doi.org/10.18653/v1/2020.acl-main.421} {On the
  cross-lingual transferability of monolingual representations}.
\newblock In \emph{Proceedings of the 58th Annual Meeting of the Association
  for Computational Linguistics}, pages 4623--4637, Online. Association for
  Computational Linguistics.

\bibitem[{Bhat et~al.(2015)Bhat, Mujadia, Tammewar, Bhat, and
  Shrivastava}]{Bhat:2014:ISS:2824864.2824872}
Irshad~Ahmad Bhat, Vandan Mujadia, Aniruddha Tammewar, Riyaz~Ahmad Bhat, and
  Manish Shrivastava. 2015.
\newblock \href {https://doi.org/10.1145/2824864.2824872} {Iiit-h system
  submission for fire2014 shared task on transliterated search}.
\newblock In \emph{Proceedings of the Forum for Information Retrieval
  Evaluation}, FIRE '14, pages 48--53, New York, NY, USA. ACM.

\bibitem[{Cao et~al.(2020)Cao, Kitaev, and Klein}]{DBLP:conf/iclr/CaoKK20}
Steven Cao, Nikita Kitaev, and Dan Klein. 2020.
\newblock \href {https://openreview.net/forum?id=r1xCMyBtPS} {Multilingual
  alignment of contextual word representations}.
\newblock In \emph{8th International Conference on Learning Representations,
  {ICLR} 2020, Addis Ababa, Ethiopia, April 26-30, 2020}. OpenReview.net.

\bibitem[{Conneau et~al.(2019)Conneau, Khandelwal, Goyal, Chaudhary, Wenzek,
  Guzm{\'a}n, Grave, Ott, Zettlemoyer, and Stoyanov}]{conneau2019unsupervised}
Alexis Conneau, Kartikay Khandelwal, Naman Goyal, Vishrav Chaudhary, Guillaume
  Wenzek, Francisco Guzm{\'a}n, Edouard Grave, Myle Ott, Luke Zettlemoyer, and
  Veselin Stoyanov. 2019.
\newblock Unsupervised cross-lingual representation learning at scale.
\newblock \emph{arXiv preprint arXiv:1911.02116}.

\bibitem[{Conneau and Lample(2019)}]{NEURIPS2019_c04c19c2}
Alexis Conneau and Guillaume Lample. 2019.
\newblock \href
  {https://proceedings.neurips.cc/paper/2019/file/c04c19c2c2474dbf5f7ac4372c5b9af1-Paper.pdf}
  {Cross-lingual language model pretraining}.
\newblock In \emph{Advances in Neural Information Processing Systems},
  volume~32, pages 7059--7069. Curran Associates, Inc.

\bibitem[{Devlin et~al.(2019)Devlin, Chang, Lee, and
  Toutanova}]{devlin-etal-2019-bert}
Jacob Devlin, Ming-Wei Chang, Kenton Lee, and Kristina Toutanova. 2019.
\newblock \href {https://doi.org/10.18653/v1/N19-1423} {{BERT}: Pre-training of
  deep bidirectional transformers for language understanding}.
\newblock In \emph{Proceedings of the 2019 Conference of the North {A}merican
  Chapter of the Association for Computational Linguistics: Human Language
  Technologies, Volume 1 (Long and Short Papers)}, pages 4171--4186,
  Minneapolis, Minnesota. Association for Computational Linguistics.

\bibitem[{Dryer and Haspelmath(2013)}]{wals}
Matthew~S. Dryer and Martin Haspelmath, editors. 2013.
\newblock \href {https://wals.info/} {\emph{WALS Online}}.
\newblock Max Planck Institute for Evolutionary Anthropology, Leipzig.

\bibitem[{Gao et~al.(2020)Gao, Wang, and Neubig}]{gao-etal-2020-improving}
Luyu Gao, Xinyi Wang, and Graham Neubig. 2020.
\newblock \href {https://doi.org/10.18653/v1/2020.findings-emnlp.319}
  {Improving target-side lexical transfer in multilingual neural machine
  translation}.
\newblock In \emph{Findings of the Association for Computational Linguistics:
  EMNLP 2020}, pages 3560--3566, Online. Association for Computational
  Linguistics.

\bibitem[{Goyal and Durrett(2019)}]{goyal-durrett-2019-embedding}
Tanya Goyal and Greg Durrett. 2019.
\newblock \href {https://doi.org/10.18653/v1/P19-1433} {Embedding time
  expressions for deep temporal ordering models}.
\newblock In \emph{Proceedings of the 57th Annual Meeting of the Association
  for Computational Linguistics}, pages 4400--4406, Florence, Italy.
  Association for Computational Linguistics.

\bibitem[{Goyal et~al.(2020)Goyal, Kumar, and
  Sharma}]{goyal-etal-2020-efficient}
Vikrant Goyal, Sourav Kumar, and Dipti~Misra Sharma. 2020.
\newblock \href {https://doi.org/10.18653/v1/2020.acl-srw.22} {Efficient neural
  machine translation for low-resource languages via exploiting related
  languages}.
\newblock In \emph{Proceedings of the 58th Annual Meeting of the Association
  for Computational Linguistics: Student Research Workshop}, pages 162--168,
  Online. Association for Computational Linguistics.

\bibitem[{Hu et~al.(2020)Hu, Ruder, Siddhant, Neubig, Firat, and
  Johnson}]{hu2020xtreme}
Junjie Hu, Sebastian Ruder, Aditya Siddhant, Graham Neubig, Orhan Firat, and
  Melvin Johnson. 2020.
\newblock \href {http://arxiv.org/abs/2003.11080} {Xtreme: A massively
  multilingual multi-task benchmark for evaluating cross-lingual
  generalization}.
\newblock \emph{CoRR}, abs/2003.11080.

\bibitem[{Kakwani et~al.(2020)Kakwani, Kunchukuttan, Golla, Gokul,
  Bhattacharyya, Khapra, and Kumar}]{kakwani2020inlpsuite}
Divyanshu Kakwani, Anoop Kunchukuttan, Satish Golla, NC~Gokul, Avik
  Bhattacharyya, Mitesh~M Khapra, and Pratyush Kumar. 2020.
\newblock inlpsuite: Monolingual corpora, evaluation benchmarks and pre-trained
  multilingual language models for indian languages.
\newblock In \emph{Proceedings of EMNLP 2020}.

\bibitem[{Khanuja et~al.(2021)Khanuja, Bansal, Mehtani, Khosla, Dey, Gopalan,
  Margam, Aggarwal, Nagipogu, Dave, Gupta, Gali, Subramanian, and
  Talukdar}]{khanuja2021muril}
Simran Khanuja, Diksha Bansal, Sarvesh Mehtani, Savya Khosla, Atreyee Dey,
  Balaji Gopalan, Dilip~Kumar Margam, Pooja Aggarwal, Rajiv~Teja Nagipogu,
  Shachi Dave, Shruti Gupta, Subhash Chandra~Bose Gali, Vish Subramanian, and
  Partha Talukdar. 2021.
\newblock Muril: Multilingual representations for indian languages.
\newblock \emph{arXiv preprint arXiv:2103.10730}.

\bibitem[{Kunchukuttan and Bhattacharyya(2020)}]{kunchukuttan2020utilizing}
Anoop Kunchukuttan and Pushpak Bhattacharyya. 2020.
\newblock \href {http://arxiv.org/abs/2003.08925} {Utilizing language
  relatedness to improve machine translation: A case study on languages of the
  indian subcontinent}.

\bibitem[{Lan et~al.(2020)Lan, Chen, Goodman, Gimpel, Sharma, and
  Soricut}]{lan2020albert}
Zhenzhong Lan, Mingda Chen, Sebastian Goodman, Kevin Gimpel, Piyush Sharma, and
  Radu Soricut. 2020.
\newblock Albert: A lite bert for self-supervised learning of language
  representations.

\bibitem[{Lauscher et~al.(2020)Lauscher, Ravishankar, Vuli{\'c}, and
  Glava{\v{s}}}]{lauscher-etal-2020-zero}
Anne Lauscher, Vinit Ravishankar, Ivan Vuli{\'c}, and Goran Glava{\v{s}}. 2020.
\newblock \href {https://doi.org/10.18653/v1/2020.emnlp-main.363} {From zero to
  hero: {O}n the limitations of zero-shot language transfer with multilingual
  {T}ransformers}.
\newblock In \emph{Proceedings of the 2020 Conference on Empirical Methods in
  Natural Language Processing (EMNLP)}, pages 4483--4499, Online. Association
  for Computational Linguistics.

\bibitem[{Littell et~al.(2017)Littell, Mortensen, Lin, Kairis, Turner, and
  Levin}]{littell2017}
Patrick Littell, David~R. Mortensen, Ke~Lin, Katherine Kairis, Carlisle Turner,
  and Lori Levin. 2017.
\newblock {URIEL} and lang2vec: Representing languages as typological,
  geographical, and phylogenetic vectors.
\newblock In \emph{Proceedings of the 15th Conference of the {E}uropean Chapter
  of the Association for Computational Linguistics: Volume 2, Short Papers}.

\bibitem[{Liu et~al.(2019)Liu, Ott, Goyal, Du, Joshi, Chen, Levy, Lewis,
  Zettlemoyer, and Stoyanov}]{liu2019roberta}
Yinhan Liu, Myle Ott, Naman Goyal, Jingfei Du, Mandar Joshi, Danqi Chen, Omer
  Levy, Mike Lewis, Luke Zettlemoyer, and Veselin Stoyanov. 2019.
\newblock Roberta: A robustly optimized bert pretraining approach.

\bibitem[{Neubig and Hu(2018)}]{neubig-hu-2018-rapid}
Graham Neubig and Junjie Hu. 2018.
\newblock \href {https://doi.org/10.18653/v1/D18-1103} {Rapid adaptation of
  neural machine translation to new languages}.
\newblock In \emph{Proceedings of the 2018 Conference on Empirical Methods in
  Natural Language Processing}, pages 875--880, Brussels, Belgium. Association
  for Computational Linguistics.

\bibitem[{Pan et~al.(2017)Pan, Zhang, May, Nothman, Knight, and Ji}]{Pan2017}
Xiaoman Pan, Boliang Zhang, Jonathan May, Joel Nothman, Kevin Knight, and Heng
  Ji. 2017.
\newblock {Cross-lingual name tagging and linking for 282 languages}.
\newblock In \emph{Proceedings of ACL 2017}, pages 1946--1958.

\bibitem[{Pfeiffer et~al.(2020{\natexlab{a}})Pfeiffer, R{\"u}ckl{\'e}, Poth,
  Kamath, Vuli{\'c}, Ruder, Cho, and Gurevych}]{pfeiffer-etal-2020-adapterhub}
Jonas Pfeiffer, Andreas R{\"u}ckl{\'e}, Clifton Poth, Aishwarya Kamath, Ivan
  Vuli{\'c}, Sebastian Ruder, Kyunghyun Cho, and Iryna Gurevych.
  2020{\natexlab{a}}.
\newblock \href {https://doi.org/10.18653/v1/2020.emnlp-demos.7}
  {{A}dapter{H}ub: A framework for adapting transformers}.
\newblock In \emph{Proceedings of the 2020 Conference on Empirical Methods in
  Natural Language Processing: System Demonstrations}, pages 46--54, Online.
  Association for Computational Linguistics.

\bibitem[{Pfeiffer et~al.(2020{\natexlab{b}})Pfeiffer, Vuli{\'c}, Gurevych, and
  Ruder}]{pfeiffer-etal-2020-mad}
Jonas Pfeiffer, Ivan Vuli{\'c}, Iryna Gurevych, and Sebastian Ruder.
  2020{\natexlab{b}}.
\newblock \href {https://doi.org/10.18653/v1/2020.emnlp-main.617} {{MAD-X}:
  {A}n {A}dapter-{B}ased {F}ramework for {M}ulti-{T}ask {C}ross-{L}ingual
  {T}ransfer}.
\newblock In \emph{Proceedings of the 2020 Conference on Empirical Methods in
  Natural Language Processing (EMNLP)}, pages 7654--7673, Online. Association
  for Computational Linguistics.

\bibitem[{Pfeiffer et~al.(2020{\natexlab{c}})Pfeiffer, Vulic, Gurevych, and
  Ruder}]{Pfeiffer2020-unks}
Jonas Pfeiffer, Ivan Vulic, Iryna Gurevych, and Sebastian Ruder.
  2020{\natexlab{c}}.
\newblock \href {http://arxiv.org/abs/2012.15562} {Unks everywhere: Adapting
  multilingual language models to new scripts}.
\newblock \emph{CoRR}, abs/2012.15562.

\bibitem[{Pires et~al.(2019)Pires, Schlinger, and
  Garrette}]{pires-etal-2019-multilingual}
Telmo Pires, Eva Schlinger, and Dan Garrette. 2019.
\newblock \href {https://doi.org/10.18653/v1/P19-1493} {How multilingual is
  multilingual {BERT}?}
\newblock In \emph{Proceedings of the 57th Annual Meeting of the Association
  for Computational Linguistics}, pages 4996--5001, Florence, Italy.
  Association for Computational Linguistics.

\bibitem[{Sardesai et~al.(2012)Sardesai, Pawar, Walawalikar, and
  Vaz}]{sardesai-etal-2012-bis}
Madhavi Sardesai, Jyoti Pawar, Shantaram Walawalikar, and Edna Vaz. 2012.
\newblock \href {https://www.aclweb.org/anthology/W12-5012} {{BIS} annotation
  standards with reference to {K}onkani language}.
\newblock In \emph{Proceedings of the 3rd Workshop on South and Southeast
  {A}sian Natural Language Processing}, pages 145--152, Mumbai, India. The
  COLING 2012 Organizing Committee.

\bibitem[{Song et~al.(2020)Song, Dabre, Mao, Cheng, Kurohashi, and
  Sumita}]{song-etal-2020-pre}
Haiyue Song, Raj Dabre, Zhuoyuan Mao, Fei Cheng, Sadao Kurohashi, and Eiichiro
  Sumita. 2020.
\newblock \href {https://doi.org/10.18653/v1/2020.acl-srw.37} {Pre-training via
  leveraging assisting languages for neural machine translation}.
\newblock In \emph{Proceedings of the 58th Annual Meeting of the Association
  for Computational Linguistics: Student Research Workshop}, pages 279--285,
  Online. Association for Computational Linguistics.

\bibitem[{{\"U}st{\"u}n et~al.(2020){\"U}st{\"u}n, Bisazza, Bouma, and van
  Noord}]{ustun-etal-2020-udapter}
Ahmet {\"U}st{\"u}n, Arianna Bisazza, Gosse Bouma, and Gertjan van Noord. 2020.
\newblock \href {https://doi.org/10.18653/v1/2020.emnlp-main.180} {{UD}apter:
  Language adaptation for truly {U}niversal {D}ependency parsing}.
\newblock In \emph{Proceedings of the 2020 Conference on Empirical Methods in
  Natural Language Processing (EMNLP)}, pages 2302--2315, Online. Association
  for Computational Linguistics.

\bibitem[{Vaswani et~al.(2017)Vaswani, Shazeer, Parmar, Uszkoreit, Jones,
  Gomez, Kaiser, and Polosukhin}]{vaswani2017attention}
Ashish Vaswani, Noam Shazeer, Niki Parmar, Jakob Uszkoreit, Llion Jones,
  Aidan~N Gomez, Lukasz Kaiser, and Illia Polosukhin. 2017.
\newblock Attention is all you need.
\newblock \emph{arXiv preprint arXiv:1706.03762}.

\bibitem[{Wang et~al.(2019)Wang, Pham, Arthur, and
  Neubig}]{wang2019multilingual}
Xinyi Wang, Hieu Pham, Philip Arthur, and Graham Neubig. 2019.
\newblock Multilingual neural machine translation with soft decoupled encoding.
\newblock \emph{arXiv preprint arXiv:1902.03499}.

\bibitem[{Wang et~al.(2020)Wang, K, Mayhew, and
  Roth}]{wang-etal-2020-extending}
Zihan Wang, Karthikeyan K, Stephen Mayhew, and Dan Roth. 2020.
\newblock \href {https://doi.org/10.18653/v1/2020.findings-emnlp.240}
  {Extending multilingual {BERT} to low-resource languages}.
\newblock In \emph{Findings of the Association for Computational Linguistics:
  EMNLP 2020}, pages 2649--2656, Online. Association for Computational
  Linguistics.

\bibitem[{Wu and Dredze(2020)}]{wu-dredze-2020-explicit}
Shijie Wu and Mark Dredze. 2020.
\newblock \href {https://doi.org/10.18653/v1/2020.emnlp-main.362} {Do explicit
  alignments robustly improve multilingual encoders?}
\newblock In \emph{Proceedings of the 2020 Conference on Empirical Methods in
  Natural Language Processing (EMNLP)}, pages 4471--4482, Online. Association
  for Computational Linguistics.

\bibitem[{Xia et~al.(2019)Xia, Kong, Anastasopoulos, and
  Neubig}]{xia-etal-2019-generalized}
Mengzhou Xia, Xiang Kong, Antonios Anastasopoulos, and Graham Neubig. 2019.
\newblock \href {https://doi.org/10.18653/v1/P19-1579} {Generalized data
  augmentation for low-resource translation}.
\newblock In \emph{Proceedings of the 57th Annual Meeting of the Association
  for Computational Linguistics}, pages 5786--5796, Florence, Italy.
  Association for Computational Linguistics.

\bibitem[{Xu et~al.(2018)Xu, Yang, Otani, and Wu}]{xu-etal-2018-unsupervised}
Ruochen Xu, Yiming Yang, Naoki Otani, and Yuexin Wu. 2018.
\newblock \href {https://doi.org/10.18653/v1/D18-1268} {Unsupervised
  cross-lingual transfer of word embedding spaces}.
\newblock In \emph{Proceedings of the 2018 Conference on Empirical Methods in
  Natural Language Processing}, pages 2465--2474, Brussels, Belgium.
  Association for Computational Linguistics.

\bibitem[{Zhang et~al.(2017)Zhang, Liu, Luan, and
  Sun}]{zhang-etal-2017-adversarial}
Meng Zhang, Yang Liu, Huanbo Luan, and Maosong Sun. 2017.
\newblock \href {https://doi.org/10.18653/v1/P17-1179} {Adversarial training
  for unsupervised bilingual lexicon induction}.
\newblock In \emph{Proceedings of the 55th Annual Meeting of the Association
  for Computational Linguistics (Volume 1: Long Papers)}, pages 1959--1970,
  Vancouver, Canada. Association for Computational Linguistics.

\end{thebibliography}

\section*{Appendix}

\subsection*{A\hspace{10pt}Additional Experiments with Contrastive Loss}
\label{sec:cstvlossexp}
Apart from MSE loss, we also experimented with the recently proposed Contrastive Loss. We present the results of using contrastive loss with various methods of dictionary lookups as described in Section 4 of the paper, in Table~\ref{tab:cstv}

\begin{table}[h]
        \centering
        \begin{adjustbox}{width=0.45\textwidth}
        \begin{tabular}{|l|l|l|l|l|}
        \hline
        Loss & Dict Lookup & NER & POS & Text C.\\
        \hline
        \multicolumn{5}{|c|}{\bf Punjabi}\\
        \hline
        cstv & first  & 73.1 &  80.7 & 75.5\\
        cstv & max  & 62.1 &  79.8 &  73.4\\
        cstv & root-weighted  & 72.1 &  78.5 &  77.9\\
        cstv &  weighted &  68.2  & 80.8  & 79.4\\
        \hline
        \hline 
        \multicolumn{5}{|c|}{\bf Gujarati}\\
        \hline
        cstv & first  & 39.9 &  83.3 &  80.4\\
        cstv & max  & 38.9  & 84.1 &  80.8\\
        cstv & root-weighted & 39.9 & 83.1 &  76.0\\
        cstv & weighted & 40.2 & 84.0 &  81.6\\
        \hline
        \hline 
        \multicolumn{5}{|c|}{\bf Bengali}\\
        \hline
        cstv & first & 56.2 &  67.7 & 77.2\\
        cstv & max  & 56.9 & 69.2 & 76.9\\
        cstv & root-weighted  & 58.5 & 71.1 &  70.9\\
        cstv & weighted & 56.6 & 67.6 & 76.5\\
        \hline
      \end{tabular}
      \end{adjustbox}
      \caption{Evaluations on NER, POS tagging and Text Classification in \our  \ using Contrastive Loss with different methods of dictionary lookup} 
      \label{tab:cstv}
\end{table}

\subsection*{B\hspace{10pt}POS Tagset mapping between Penn Treebank Tagset and BIS Tagset}
\label{sec:tagsetmapping}
For the POS experiments involving m-BERT as the base model, we fine-tune our trained model with both English and Hindi training data and calculate zero-shot results on the target language. However, the English dataset that we used was annotated using Penn Treebank Tagset while the rest of the languages were annotated using BIS Tagset. We came up with a mapping between the Penn Tags and the BIS Tags so that the English POS dataset becomes consistent with the Hindi counterpart. Table ~\ref{tab:posmap} contains the mapping that we used for the said conversion. Note that since we are using top-level tags (e.g Pronouns) instead of sub-level tags (e.g Personal Pronouns, Possessive Pronouns) for the POS classification, the mapping is also done to reflect the same.

\begin{table}[h]
\centering
\begin{adjustbox}{width=0.45\textwidth,height=0.35\textwidth}
\begin{tabular}{|c|c|c|c|}
\hline
\multicolumn{1}{|c|}{\textbf{\begin{tabular}[c]{@{}c@{}}Penn Tagset\end{tabular}}} &
  \multicolumn{1}{c|}{\textbf{BIS Tagset}} &
\multicolumn{1}{|c|}{\textbf{\begin{tabular}[c]{@{}c@{}}Penn Tagset\end{tabular}}} &
  \multicolumn{1}{c|}{\textbf{BIS Tagset}} \\ \hline
CC    & CC  & CD    & QT  \\ \hline
EX    & RD  &  FW    & RD  \\ \hline
IN    & PSP & JJ    & JJ  \\ \hline
JJR   & JJ  & JJS   & JJ  \\ \hline
LS    & QT  & MD    & V   \\ \hline
NN    & N   & NNS   & N   \\ \hline
NNP   & N   & NNPS  & N   \\ \hline
POS   & PSP & PRP   & PR  \\ \hline
PRP\$ & PR  & RB    & RB  \\ \hline
RBR   & RB  & RBS   & RB  \\ \hline
RP    & RP  & SYM   & RD  \\ \hline
TO    & RP  & UH    & RP  \\ \hline
VB    & V   & VBD   & V   \\ \hline
VBG   & V   & VBN   & V   \\ \hline
VBP   & V   & VBZ   & V   \\ \hline
WP    & PR  & WP\$  & PR  \\ \hline
AFX   & RD  & -LRB- & RD  \\ \hline
-RRB- & RD  & \begin{tabular}[c]{@{}l@{}}\# . , \$ “ (\\ ) :  - ‘’  ‘\end{tabular} & RD \\ \hline

PDT &
  \begin{tabular}[c]{@{}l@{}}all, half: QT\\ such: DM\\ "default": QT\end{tabular} &

WDT &
  {\color[HTML]{000000} \begin{tabular}[c]{@{}l@{}}which, that : PR\\ whatever: RP\\ "default": PR\end{tabular}} \\ \hline
  
 DT &
  \begin{tabular}[c]{@{}l@{}}some, every,\\ both, all,\\ another, a,\\ an: QT\\ this, these,\\ the: DM\\ those, that: PR\\ "default": QT\end{tabular} &
WRB &
  \begin{tabular}[c]{@{}l@{}}how,wherever,\\ when, where: PR\\ whenever: RB\\ why: RB\\ "default" : PR\end{tabular} \\ \hline

\end{tabular}
\end{adjustbox}
\caption{Tagset mapping between Penn Treebank and BIS. For some tags in Penn treebank (e.g. DT), we decided that a one-to-many mapping was appropriate based on a word-level division}
\label{tab:posmap}
\end{table}



\end{document}